\documentclass[10pt, conference, compsocconf]{IEEEtran}
\ifCLASSINFOpdf
\else
\fi
\usepackage{epsfig}
\usepackage{graphicx}
\usepackage{amsmath}
\usepackage{amssymb}
\usepackage{multirow}
\usepackage{subcaption}
\begin{document}
\title{Towards Accurate Marker-less Human Shape and Pose Estimation over Time}


\author{%
 Yinghao Huang$^{1}$, Federica Bogo$^{2}$, Christoph Lassner$^{3,*}$, Angjoo Kanazawa$^{4}$\\ Peter V. Gehler$^{5,*}$, Javier Romero$^{3}$, Ijaz Akhter$^{6}$, Michael J. Black$^{1}$ \\\\
 $^{1}$Max Planck Institute for Intelligent Systems, T\"ubingen, Germany\\
 $^{2}$Microsoft, $^3$Body Labs Inc., $^{4}$UC Berkeley\\ 
 $^{5}$University of W\"urzburg, $^{6}$Australian National University \\
}


%Telephone: (800) 555--1212, Fax: (888) 555--1212}


\maketitle
\let\thefootnote\relax\footnote{* The work was performed at MPI for Intelligent Systems.}

\begin{abstract}
Existing markerless motion capture methods often assume known backgrounds, static cameras, and sequence specific motion priors, limiting their application scenarios.
Here we present a fully automatic method that, given multi-view videos, estimates 3D human pose and body shape. 
We take the recently proposed SMPLify method \cite{bogo2016keep} as the base method and extend it in several ways. 
First we fit a 3D human body model to 2D features detected in multi-view images. 
Second, we use a CNN method to segment the person in each image and fit the 3D body model to the contours, further improving accuracy. 
Third we utilize a generic and robust DCT temporal prior to handle the left and right side swapping issue sometimes introduced by the 2D pose estimator. 
Validation on standard benchmarks shows our results are comparable to the state of the art and also provide a realistic 3D shape avatar. 
We also demonstrate accurate results on HumanEva and on challenging monocular sequences of dancing from YouTube.
\end{abstract}


\begin{figure*}
\begin{center}
        \scalebox{1.0}[1.0]{\includegraphics[width=0.24\linewidth]{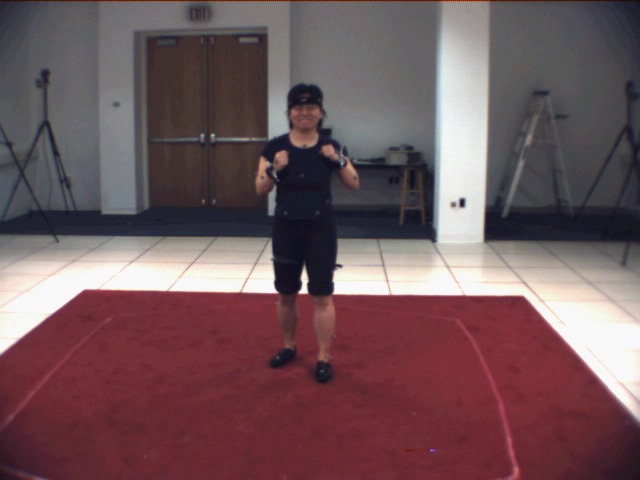}}        
        \scalebox{1.0}[1.0]{\includegraphics[width=0.24\linewidth]{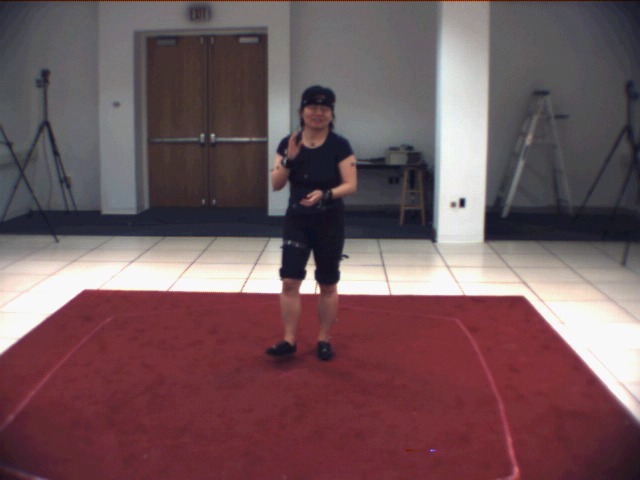}} 
        \scalebox{1.0}[1.0]{\includegraphics[width=0.24\linewidth]{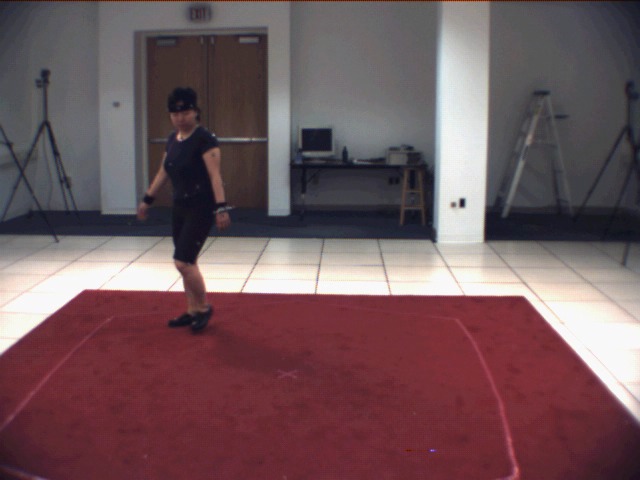}}
        \scalebox{1.0}[1.0]{\includegraphics[width=0.24\linewidth]{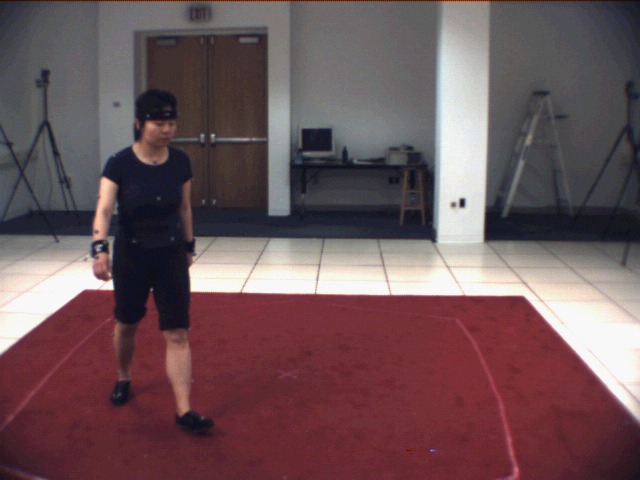}}         \\
        %\vspace{-0.25cm}   
        
		\scalebox{1.0}[1.0]{\includegraphics[width=0.24\linewidth]{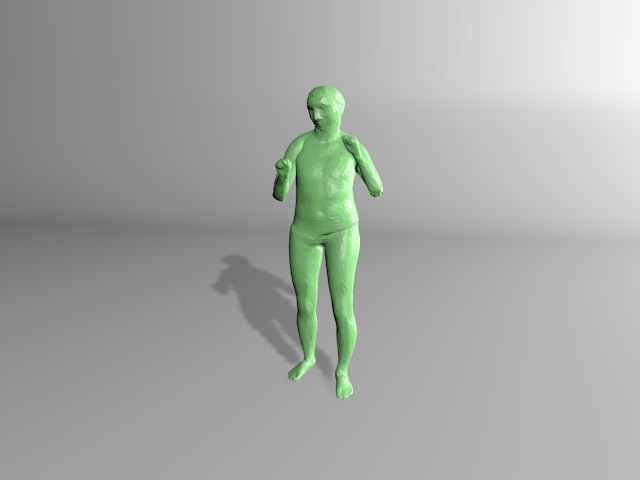}}        
        \scalebox{1.0}[1.0]{\includegraphics[width=0.24\linewidth]{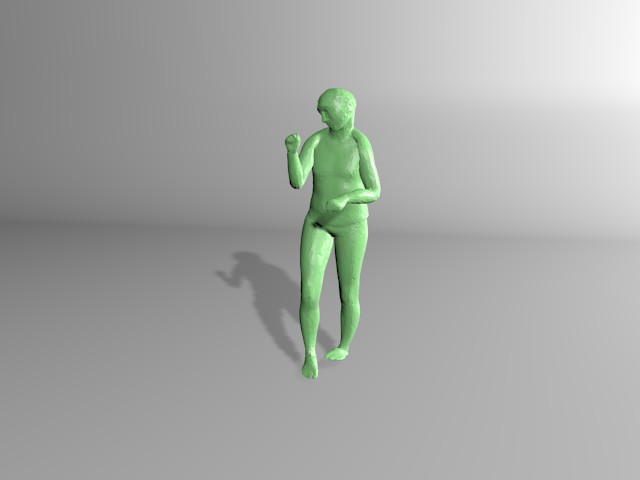}} 
        \scalebox{1.0}[1.0]{\includegraphics[width=0.24\linewidth]{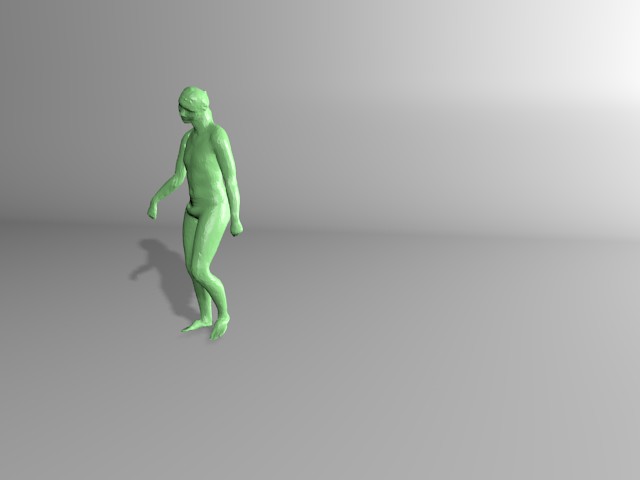}}
        \scalebox{1.0}[1.0]{\includegraphics[width=0.24\linewidth]{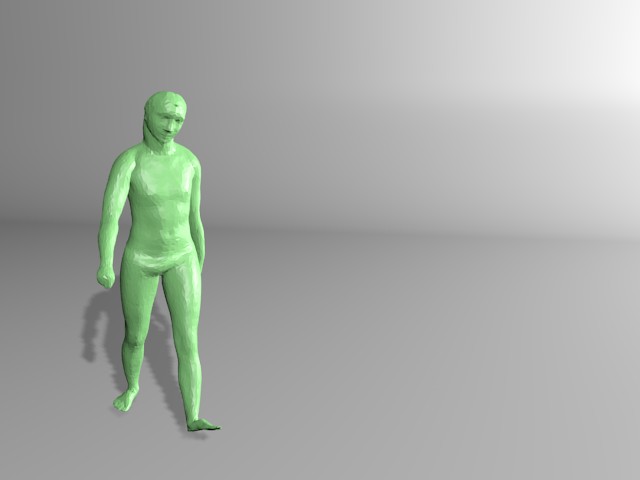}}     \\

        \scalebox{1.0}[1.0]{\includegraphics[width=0.24\linewidth]{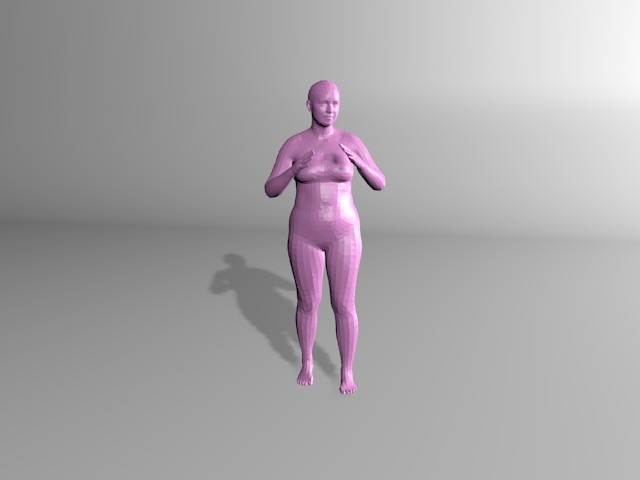}}        
        \scalebox{1.0}[1.0]{\includegraphics[width=0.24\linewidth]{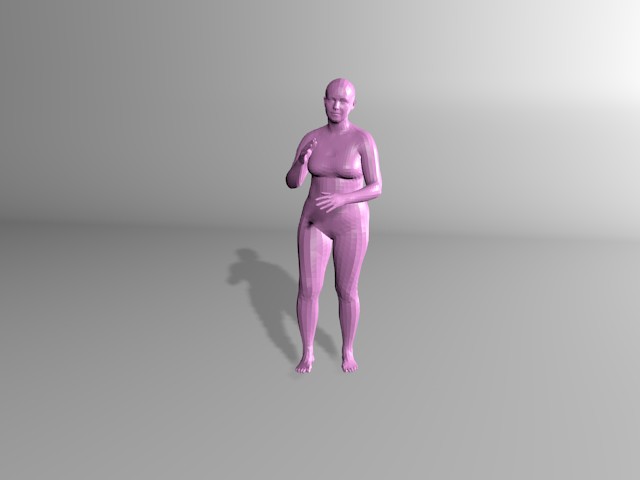}} 
        %\scalebox{1.0}[1.0]{\includegraphics[width=0.19\linewidth]{Figure/5_1383_vis.jpg}}
        \scalebox{1.0}[1.0]{\includegraphics[width=0.24\linewidth]{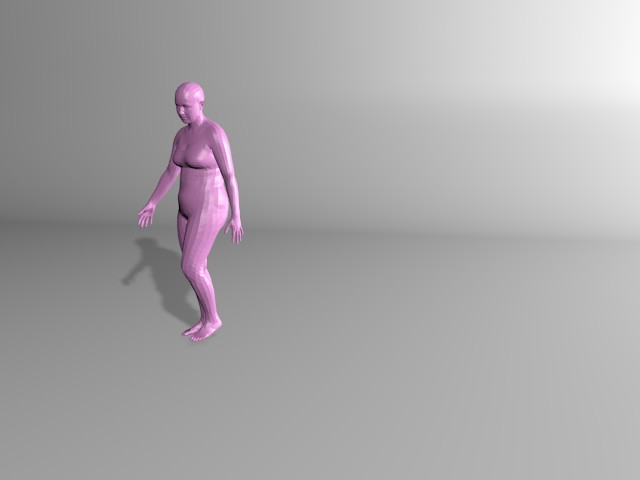}}
        \scalebox{1.0}[1.0]{\includegraphics[width=0.24\linewidth]{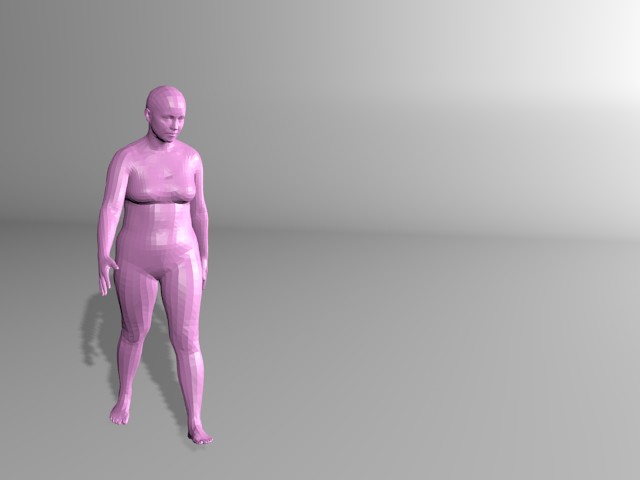}}  
    \captionof{figure}{Given multi-view videos, our method can not only yield more accurate 3D pose estimation results, but also more realistic and natural meshes than the state of the art. The entire process is fully automatic. From up to bottom: example input frames; meshes returned by \cite{rhodin2016general}; meshes generated by our method.}
    \centering
\label{fig:abstract}
\end{center}%
\end{figure*}

% \fi
\IEEEpeerreviewmaketitle

\section{Introduction}
The markerless capture of human motion (mocap) has been a long term goal of the community. 
While there have been many proposed approaches and even commercial ventures, existing methods typically operate under restricted environments.
Most commonly, such methods exploit background ``subtraction,'' assuming a known and static background,
and the most accurate methods employ strong prior assumptions about the motion of the actor.
In many cases, the best results on benchmarks like HumanEva \cite{sigal2010humaneva} are obtained by training on the same motion by the same actor as is evaluated at test time \cite{amin2013multi}.
Here we provide a solution for markerless mocap that is more accurate than the recent state of the art but is also less restrictive.

There are four key components to our approach. 
First our approach exploits SMPL \cite{loper2015smpl},  a realistic, low-dimensional, 3D parametric model of the human body.
Second we use a Convolutional Neural Network (CNN) to compute putative 2D joint locations in multiple camera images. 
We  then fit the 3D parametric model to the 2D joints robustly. 
This extends the SMPLify approach for pose and shape estimation \cite{bogo2016keep} from a single image to multi-camera data. 

Third, we go beyond SMPLify \cite{bogo2016keep}  to use a deep CNN to also segment people from images \cite{LassnerCVPR2017}. 
This removes the need for a background image and makes the approach more general. 
We fit our 3D body model to both the 2D joints and the estimated silhouettes and show that the silhouettes provide significantly improved accuracy and realism to the mocap. 

Since 2D joints estimated by CNNs sometimes confuse left and right parts of the body, the image evidence alone is not enough for a reliable 3D solution.
Consequently we exploit temporal information to resolve such errors. 
This leads to the fourth component in which we exploit a generic temporal prior based on the insight that human motions can be captured by a low-dimensional Discrete Cosine Transform (DCT) basis \cite{akhter2012bilinear}.
%a more advanced version is presented in \cite{akhter2012bilinear}. 
We implement this DCT temporal term robustly and show that it improves performance yet requires no training data.

We call the method MuVS  (Multi-View SMPLify) and evaluate it quantitatively on HumanEva \cite{sigal2010humaneva} and Human3.6M \cite{ionescu2014human3}. 
We find that MuVS gives an error comparable with any published result and more realistic meshes (see Figure \ref{fig:abstract}), while having fewer restrictions. 
We evaluate the method with an ablation study on HumanEva to determine which design decisions are most important. 

Additionally, our approach also works in the monocular camera setting.
We find that the temporal coherence term enables reasonable reconstruction of pose from monocular video even with a moving camera, complex background, and challenging motions.
We evaluate this quantitatively on HumanEva \cite{sigal2010humaneva} and some challenging dancing video sequences from Youtube. 
The software will be made available for research purposes.

\section{Related Work}
The majority of previous works only handle one aspect of the two closely related problems: markerless 3D human body shape and pose estimation. 
Some of these target 3D pose estimation \cite{amin2013multi, deutscher2005articulated, du2016marker,  gall2010optimization, ramakrishna2012reconstructing,  sigal2012loose, yao2011learning}. 
They formulate it as a discriminative problem, directly inferring 3D pose from 2D image features, assuming no explicit human body model. 
Amin et al.~\cite{amin2013multi} extend single-view based pictorial structure to multi-view cases, jointly infer 2D joint location of all views, then use linear-triangulation to obtain the 3D joints. 
Yao et al.~\cite{yao2011learning} propose a stochastic gradient-based method for a Gaussian Process Latent Variable Model (GPLVM), which shows good optimization properties. 
Uncertainty over estimated 2D image features has also been considered. 
Zhou et al.~\cite{zhou2016sparseness} introduce sparsity prior over human pose, and jointly handle the pose and 2D location uncertainty, while Kazemi et al.~\cite{kazemi2013multi} address the body part correspondence problem by optimizing latent variables.
Similar ideas are proposed by Simo-Serra et al.~\cite{simo2013joint}, in which they also estimate 2D and 3D pose at the same time.
Twin Gaussian processes \cite{bo2010twin} have also been used on this problem. 
Most recently deep learning methods achieve the most accurate pose estimation results \cite{du2016marker, moreno20163d, popa2017deep, tekin2015direct, trumble2016deep}. To address their need for huge amounts of training data, Yasin et al.~\cite{yasin2016dual} propose a dual-source approach. 
Pavlakos et al.~\cite{pavlakos2017volumetric} directly regress 3D pose from RGB image via CNNs in a coarse-to-fine manner. 

The second major set of approaches use an explicit intermediate human body representation, which effectively assists pose estimation but often lacks realism \cite{belagiannis20143d, deutscher2000articulated,  sigal2012loose, stoll2011fast}. 
Common human body representations include the Articulated Human Body Model \cite{deutscher2005articulated}, 3D Pictorial Structures \cite{belagiannis20143d,sigal2012loose}, the sum-of-Gaussians model \cite{stoll2011fast}, and the Triangulated Mesh Model \cite{sigal2007combined}.
These models are usually utilized to represent the structure of the human body, thus facilitating the inference of pose parameters. 
Sometimes the body mesh is also considered, but in an abstract or coarse way, without consideration of the shape details.  

Estimating both the pose and surface mesh, usually requires complex global optimization \cite{gall2010optimization, gall2009motion}.
Often the  silhouette of the body is assumed to be known \cite{bualan2008naked} and manual initialization or a pre-scanned surface mesh is required \cite{ahmed2005automatic, ballan2008marker, de2008performance,  hasler2010multilinear, ilic2006implicit, jain2010moviereshape, plankers2003articulated, starck2003model, wu2012full, vlasic2008articulated}. 
Balan et al.~\cite{balan2007detailed} address this problem by fitting a SCAPE body model \cite{anguelov2005scape} to multi-view silhouettes.
Their initialization method is complex and they do not integrate information over time. 
Another very recent work, concurrent with ours, is the one proposed in \cite{pavlakos2017harvesting}. 
They also use CNNs to detect 2D joints, then fit a 3D pictorial structures model to the detections. 
Their method only returns 3D joints as output, while ours estimates body shape and pose together. 
The method proposed in \cite{VNect_SIGGRAPH2017} simultaneously regresses 2D and 3D joints from monocular video via one CNN, then fits a skeleton model to the 3D joint estimations, achieving real-time performance.

The most similar recent work addresses fully automatic estimation of 3D pose and shape from monocular images \cite{bogo2016keep} and multi-views videos \cite{rhodin2016general}.
The SMPLify algorithm proposed by Bogo et al.~\cite{bogo2016keep} makes it possible to simultaneously obtain 3D pose and convincing body shape from a single image, without requiring any user intervention, and without assuming background extraction or complex optimization techniques. 
Based on the state-of-the-art 3D human body model, SMPL \cite{loper2015smpl}, they infer human shape and pose parameters by fitting the projection of 3D SMPL joints to 2D joints estimated via a 2D joint detector like DeepCut or CPM \cite{pishchulin16cvpr, wei2016convolutional}. 
Ambiguity issues are handled by applying priors learned from the large-scale public CMU dataset \cite{cmudataset}, which is vital for their method to yield valid results. 
Rhodin et al.~\cite{rhodin2016general} propose a method that works on multi-view videos. Built upon a sum-of-Gaussian shape model \cite{rhodin2015versatile, stoll2011fast}, their algorithm first initializes the pose of each Gaussian blob, then refines the pose and shape of each blob via the body contour approximation with image gradients. 
As in Bogo et al.~\cite{bogo2016keep}, they use deep learning to estimate 2D joints to get rough joint locations in each view. 
Also they enforce temporal coherence by penalizing acceleration between frames.

%Another related work is the one proposed in \cite{xu2017monoperfcap}, where general human performance capture from monocular video is addressed.

The general performance capture method in \cite{xu2017monoperfcap} is also closely related to ours, and works on monocular videos. 
A specific mesh and skeleton for each actor is required in advance and sometimes manual labour is needed. 
In contrast, our method runs fully automatically and determines the skeleton configuration and surface mesh together with the pose in the process.

We go beyond SMPLify by extending it to multi-view and monocular videos in a principled way. We show that there are important additional cues other than 2D joints to utilize, like silhouettes and temporal coherence. 
Though conceptually similar to the method in \cite{rhodin2016general}, our framework differs in important respects.
First, we use explicit segmentation to obtain the body contour. 
Second, we use a DCT basis as the temporal prior model. 
As a general temporal smoothness model, DCT can be applied in any video sequence, without the need of learning from a training dataset.
Third, in contrast to the sum-of-Gaussian model \cite{rhodin2015versatile}, we use the SMPL \cite{loper2015smpl} body model, which naturally encodes the statistical shape and pose dependency between different body parts in a holistic way.
This enables our method to, not only estimate accurate 3D joint locations, but also a realistic body mesh. 
This facilitates future modification and animation. In comparison, a volumetric skinning approach is utilized in \cite{rhodin2016general} to estimate the actor body surface from the Gaussian representation. Their surface is coarser and does not allow for detailed deformations.
Finally, we demonstrate that our method can be applied on monocular videos, unlike the method in \cite{rhodin2016general}.

\section{2D Joints and Contour Segmentation}
\begin{figure}
\begin{center}
    \begin{subfigure}{0.23\textwidth}
        \scalebox{1.0}[1.0]{\includegraphics[width=\linewidth]{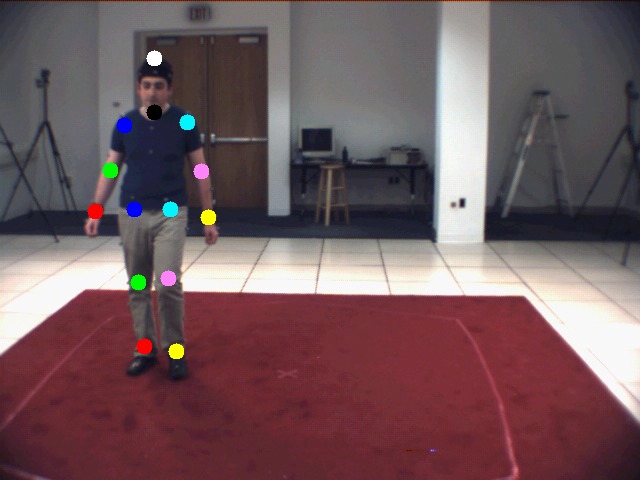}}
        %\caption{}        
        \caption{Estimated 2D joints}        
    \end{subfigure}      
    \begin{subfigure}{0.23\textwidth}
        \scalebox{1.0}[1.0]{\includegraphics[width=\linewidth]{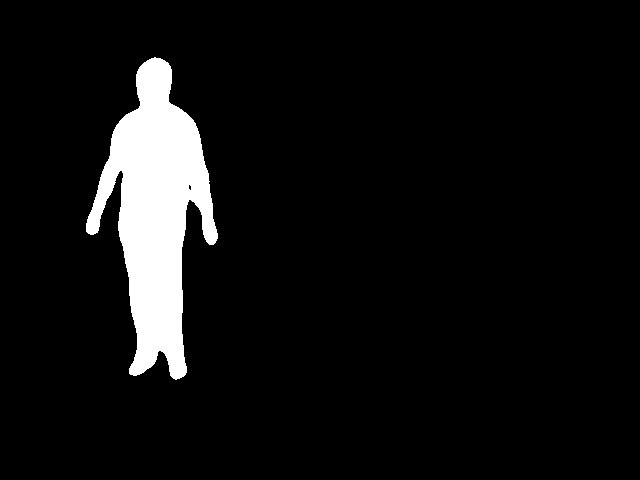}}
        %\caption{}        
        \caption{Segmented silhouette}        
    \end{subfigure}          
\caption{Automatically estimated 2D joint locations using DeepCut \cite{pishchulin16cvpr}  and the silhouette estimated via \cite{LassnerCVPR2017};
here shown on the HumanEva dataset \cite{sigal2010humaneva}.}
\label{fig:deepcut}
\end{center}
\end{figure}

Our method takes as input a set of 2D body joints and segmentation of the body from the background.
For a direct quantitative comparison with SMPLify on standard test datasets, we use the same CNN-based joint estimator, DeepCut \cite{pishchulin16cvpr}.
For more complex videos from the Internet, we use the real-time version of the CPM method \cite{CaoCVPR2017} since we find it is more reliable than DeepCut.
We also use a CNN trained to estimate human segmentations \cite{LassnerCVPR2017}.
Both of these are fully automatic and computed by CNNs \cite{pishchulin16cvpr,wei2016convolutional} trained on generic databases, which do not overlap with any of our test data.
Illustrative joint estimation and human body segmentation results are shown in Figure \ref{fig:deepcut}. 

\section{Multi-view SMPLify}

Here we first extend SMPLify to multiple camera views, then further extend it over time.  
Given the 2D joints and silhouettes for all the input frames for each camera view, we estimate the 3D pose for each time instant.
We then combine information from all the views to estimate a consistent 3D human shape over time.
Consequently, our algorithm is composed of two consecutive stages described in detail below.

In the first stage, a separate SMPL model is fit to all views independently at each time instant.
The extension of the public SMPLify code to multiple views is straightforward: we estimate the shape and pose using information from all camera views.
This gives a fully automatic approach to multi-camera marker-less motion capture.
In the case of the original single-view SMPLify, the 3D pose and shape may be quite ambiguous given 2D joints and the method relied heavily on priors to prevent interpenetration.
In contrast, with as few as 2 views, many of these ambiguities go away.
After that, the silhouette is used to refine the estimated shape, which is then more faithful to the observed body.

In the second stage, we first estimate a consistent 3D shape across the entire sequence by taking the median of all the shape parameters obtained in the first stage.
The pose parameters for each frame are initialized with their values from the first stage. 
We then consider a set of consecutive frames together and regularize the motion in time.
We do this by minimizing the projected joint error while encouraging the trajectory of each 3D joint to be well represented by a set of low-d DCT basis vectors \cite{akhter2012bilinear}. 
This temporal smoothing helps remove errors caused by inaccurate 2D joint estimates, which may be noisy and contain errors.
In particular CNNs sometimes detect spurious points or suffer from left/right ambiguity. 

\subsection{Stage One: Per-frame Fitting}
As in SMPLify, we use SMPL as our underlying shape representation. SMPL is a state-of-the-art statistical human body model \cite{loper2015smpl}, which is controlled by two sets of parameters, one for body shape, the other for pose. 
More formally, SMPL is defined as $M(\boldsymbol{\beta}, \boldsymbol{\theta}; \Phi)$, where $\boldsymbol{\beta}$ is a vector of shape parameters that are responsible for the 3D body shape due to identity, and $\boldsymbol{\theta}$ is a vector of pose parameters representing body part rotations in a kinematic tree.
The fixed parameters $\Phi$ are learned from a large number of 3D body meshes. For the detailed meaning of all these parameters, we refer the reader to \cite{loper2015smpl}.

We first estimate the shape and pose parameters of the SMPL model for each time instant. Given the corresponding 2D joint estimates $\{J_{est}^1, J_{est}^2, \dots, J_{est}^{|V|}\}$ for all the different views $V$, we formulate the energy function as the following:
\begin{equation}
E_M(\boldsymbol{\beta}, \boldsymbol{\theta}) = E_P(\boldsymbol{\beta}, \boldsymbol{\theta}) + \\ \sum_{v =1}^{V} E_J(\boldsymbol{\beta}, \boldsymbol{\theta}; K_v, J^v_{\mathrm{est}}) \,,
\end{equation}
where $E_p$ is the prior term, $K_v$ are the camera parameters for view $v$, and $E_J$ is the joint fitting term (i.e.~the data term). 
Note that here we remove the other priors used in SMPLify, because in multi-view cases the solution is better constrained. $E_P$ is composed of two terms: a shape prior $E_\beta$ and a pose prior $E_\theta$. The pose prior is learned from the CMU dataset \cite{cmudataset}, while the shape prior is leaned from the SMPL body shape training data.
\begin{equation}
    E_P(\boldsymbol{\beta}, \boldsymbol{\theta}) = \lambda_\theta E_\theta(\boldsymbol{\theta}) + \lambda_\beta E_\beta(\boldsymbol{\beta}) .
\end{equation}
The joint fitting term is formulated as follows:
\begin{eqnarray}
 \lefteqn{ E_J(\boldsymbol{\beta}, \boldsymbol{\theta}; K_v, J^v_{\mathrm{est}}) = }\nonumber \\
& & \sum_{\mathrm{joint} \, i} w_i \rho_{\sigma_1}(\mathit{\Pi}_{K_v}(R_\theta(J_i(\boldsymbol{\beta}))) - J^v_{\mathrm{est}, i}) \,,
\end{eqnarray}
where $J(\cdot)$ is the joint estimation function, which returns joint locations, $R$ is the rotation function, $\mathit{\Pi}$ the projection function, $w_i$ the confidence yielded by the 2D joint detection CNN. Considering the inevitable detection noise and errors in the entire process, instead of the standard squared error we use a robust Geman-McClure error function, which is defined by: 
\begin{equation}
\label{equ:rho}
    \rho_{\sigma}(e) = \frac{e^2}{\sigma^2 + e^2} \,,
\end{equation}
here $e$ is the residual error, and $\sigma$ is the robustness constant carefully chosen. 

After obtaining the initial pose and shape estimation via fitting SMPL to 2D joints, we further refine it by adding silhouette information. The fitting error between the contour rendered from the SMPL model and the CNN-segmented one is defined as:
\begin{eqnarray}
 \lefteqn{   E_S(\boldsymbol{\beta}, \boldsymbol{\theta}; K_v, U_v) = }\nonumber\\
& &\sum_{x \in \hat{S}(\boldsymbol{\beta}, \boldsymbol{\theta})} l(x,U_v)^2 + \sum_{x \in U_v} l(x,\hat{S}(\boldsymbol{\theta}, \boldsymbol{\beta}))    \,,
\end{eqnarray}
where $l(x,S)$ denotes the absolute distance from a point $x$ to a silhouette $S$;
the distance is zero when the point is inside $S$.
The first term computes the distance from points on the projected model $\hat{S}(\boldsymbol{\beta}, \boldsymbol{\theta})$ to the estimated silhouette $U_v$ for the $v$-th view,
while the second term computes the distance from points in the estimated silhouette $U_v$ to the model $\hat{S}(\boldsymbol{\beta}, \boldsymbol{\theta})$.
As in \cite{LassnerCVPR2017}, an $L_1$ distance metric is used in the second term to make it more robust to noise, while the first term uses the common $L_2$ distance. Combined with the 2D joint fitting term, the final energy function is:
\begin{equation}
    E_1(\boldsymbol{\beta}, \boldsymbol{\theta}) = E_M(\boldsymbol{\beta}, \boldsymbol{\theta}) + \sum_{v \in V}E_S(\boldsymbol{\beta}, \boldsymbol{\theta}; K_v, U_v) \,,           
\end{equation}
We found faster convergence to better solutions was obtained using a hierarchical optimization strategy: 
firstly fitting SMPL to 2D joints can yield a coarse estimation of pose and shape parameters efficiently, then adding the silhouette fitting term can further improve accuracy.

\subsection{Stage Two: Temporal Fitting} 
\begin{figure}
\begin{center}
    \begin{subfigure}[b]{0.15\textwidth}
        %\scalebox{1.0}[1.0]{\includegraphics[trim=0 0 10cm 0, clip=true, width=\linewidth]{Figure/3_frame0159_pose.jpg}}
        \scalebox{1.0}[1.0]{\includegraphics[trim=0 0 10cm 0, clip=true, width=\linewidth]{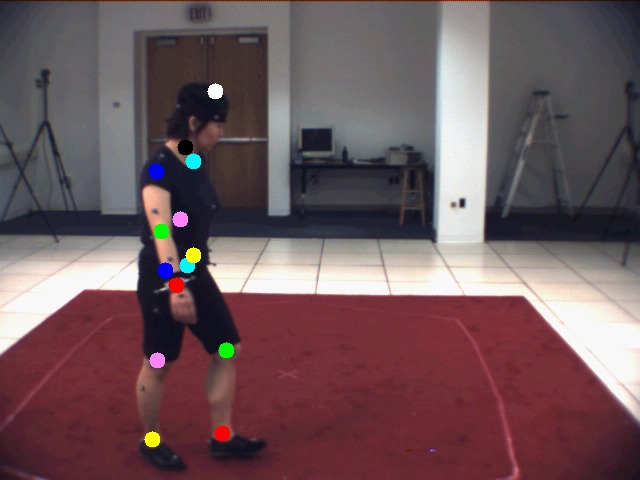}}        
        \caption{}
    \end{subfigure}
    \begin{subfigure}[b]{0.15\textwidth}
        %\scalebox{1.0}[1.0]{\includegraphics[trim=0 0 10cm 0, clip=true, width=\linewidth]{Figure/3_frame0159_res_1.jpg}}
        \scalebox{1.0}[1.0]{\includegraphics[trim=0 0 10cm 0, clip=true, width=\linewidth]{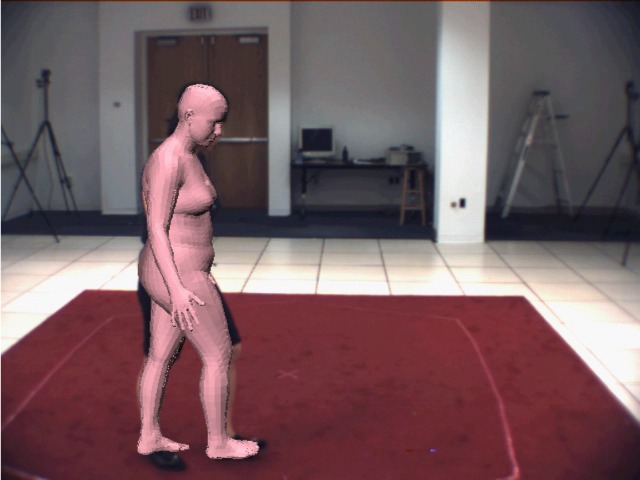}}        
        \caption{}
    \end{subfigure}
    \begin{subfigure}[b]{0.15\textwidth}
        %\scalebox{1.0}[1.0]{\includegraphics[trim=0 0 10cm 0, clip=true, width=\linewidth]{Figure/3_frame0159_res_2.jpg}}
        \scalebox{1.0}[1.0]{\includegraphics[trim=0 0 10cm 0, clip=true, width=\linewidth]{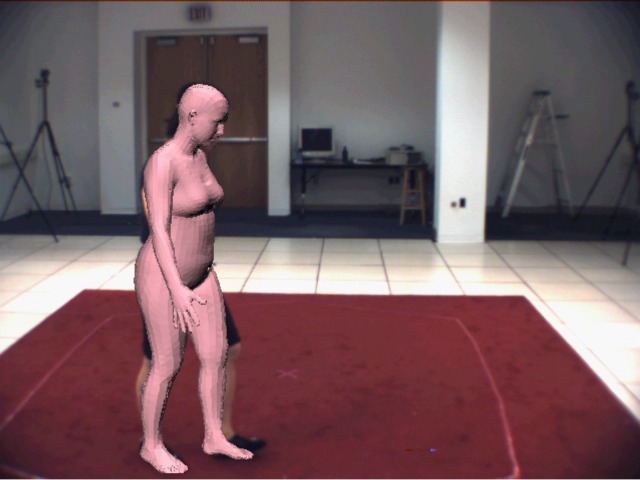}}        
        \caption{}
    \end{subfigure}
\caption{DCT based temporal prior helps alleviate the leg swap problem. a): Pose detection with leg swap; b): MuVS without DCT prior; c): MuVS with DCT prior.}
\label{fig:swap}
\end{center}
\end{figure}

One obvious shortcoming of the algorithm used in the first stage is that it does not take into account the temporal relationship between consecutive frames, while in real life human motions usually present consistency. What is more, due to the lack of texture, occlusion, similarity to the background and other noise, the joint  estimator can be erroneous in ambiguous cases. One of these errors is leg swap, which is demonstrated in Figure \ref{fig:swap}. Sometimes these errors can be difficult to automatically correct in single frame settings. 
By processing several consecutive frames simultaneously, we can greatly alleviate these types of errors. 

To make our algorithm more efficient, in this stage, we do not consider the silhouette, and only use 2D joints. 
The silhouette's value is in estimating the body shape in the first stage.
We study the effect of this choice in our ablation study. Using the obtained median shape $\boldsymbol{\hat{\beta}}$ and pose parameters $\boldsymbol{\Theta}$ from the first stage, we optimize the following objective, which is composed of the 2D joint fitting term and low-dimensional DCT reconstruction term $\boldsymbol{B}$ with corresponding coefficients $\boldsymbol{C}$: 
\begin{multline}
    E_2(\boldsymbol{\Theta}, \boldsymbol{C}; \boldsymbol{\hat{\beta}}, N) = \sum_{n = 1}^N E_M(\boldsymbol{\hat{\beta}}, \boldsymbol{\theta}_n) + \\
    \sum_{\mathrm{joint} \, e} \sum_{d \in \{X, Y, Z\}} \lambda_T E_T(\boldsymbol{C}_{e,d}, \boldsymbol{D}_{e,d}; \boldsymbol{\hat{\beta}}, \boldsymbol{B}, N),    
\end{multline}
here $\boldsymbol{\Theta} = \{\boldsymbol{\theta}_1, \boldsymbol{\theta}_2, \dots, \boldsymbol{\theta}_N\}$ is the set of pose parameters for the $N$ frames, $\boldsymbol{C}$ are the corresponding DCT coefficients, $\boldsymbol{D}$ is the collection of all 3D SMPL joints across these frames, while $\boldsymbol{D}_{e,d}$ represents the vector constructed from $d$-coordinate of the $e$-th SMPL joints, which is defined as: 
\begin{equation*}
\boldsymbol{D}_{e,d} = [R_{\theta_1}(J_d(\boldsymbol{\hat{\beta}}))_e, R_{\theta_2}(J_d(\boldsymbol{\hat{\beta}}))_e, \dots, R_{\theta_N}(J_d(\boldsymbol{\hat{\beta}}))_e]
\end{equation*}
where $e \in \{1, 2, \dots, N\}$ and $d \in \{X, Y, Z\}$. 
We encourage the trajectory $\boldsymbol{D}_{e,d}$ across $N$ frames to be well approximated by some low-dimensional DCT basis $\boldsymbol{B}$:  
\begin{equation}
    E_T(\boldsymbol{c}, \boldsymbol{d}; \boldsymbol{\beta}, \boldsymbol{B}, N) = \sum_{j=1}^N \rho_{\sigma_2}(\boldsymbol{d}_j - (\boldsymbol{B}\boldsymbol{c})_j),
\end{equation}
where $\rho$ is the same function introduced in Eq. \ref{equ:rho}.
Note that the temporal smoothness prior is formulated on the 3D SMPL joint locations.

\begin{table*}
\begin{center}
\begin{tabular}{|l c c c c c c c c|}
\hline
       & \multicolumn{3}{c}{Walking} & \multicolumn{3}{c}{Boxing} & Mean & Median\\
Method & S1 & S2 & S3 & S1 & S2 & S3 & &\\
\hline
MuVS\textsuperscript{2} & 59.22 & 66.81 & 88.60 & 79.51 & 78.68 & 88.34 & 76.86 & 79.10\\
MuVS\textsuperscript{2, S} & 54.35 & 56.06 & 80.95 & 70.27 & 72.01 & 79.01 & 68.78 & 71.14 \\
MuVS\textsuperscript{2, S, T} & 50.14 & 56.11 & 79.55 & 68.96 & 71.73 & 78.45 & 67.49 & 70.35\\
MuVS\textsuperscript{2, S, T, H} & 39.28 & 45.81 & 64.63 & 55.12 & 56.49 & 57.09 & 53.07 & 55.81 \\
MuVS\textsuperscript3 & 52.50 & 62.76 & 82.51 & 72.86 & 73.10 & 80.42 & 70.69 & 72.98\\
MuVS\textsuperscript{3, S} & 47.21 & 52.72 & 75.04 & 64.88 & 68.39 & 71.98 & 63.37 & 66.64 \\
MuVS\textsuperscript{3, S, T} & 43.11 & 53.37 & 73.56 & 64.00 & 67.94 & 71.44 & 62.23 & 65.97 \\
MuVS\textsuperscript{3, S, T, H} & \textbf{35.51} & \textbf{44.22} & \textbf{61.30} & \textbf{49.67} & \textbf{53.89} & \textbf{51.37} & \textbf{49.33} & \textbf{50.52} \\
\hline
\end{tabular}
\end{center}
\caption{Ablation results on HumanEva. 3D joint errors in {\it mm}. Here labels 2/3 mean using the first 2/3 camera views; S means silhouette fitting term; T means temporal fitting term; and H means adding silhouette fitting term at the second stage. The same notation is used in the rest of the paper.}
\label{tab:ablation}
\end{table*}

\subsection{Implementation Details}
We implement our entire algorithm in Python. The two involved optimization problems are conducted using Powell's dogleg method \cite{nocedal2006numerical},  OpenDR \cite{loper2014opendr} and Chumpy \cite{Chumpy}. In the first stage, all the parameters to optimize are initialized in the same way as \cite{bogo2016keep}. For the second stage, we choose 30 consecutive frames as a unit, and use the first 10 DCT components to act as the bases $\boldsymbol{B}$. For 4 views with 500x500 images, on a normal PC with 12GB RAM and 4 cores, the first stage of our method usually takes around 70 seconds for each frame, while each temporal unit in the second stage takes around 12 minutes. All the weights are empirically chosen by running our method on the training dataset of HumanEva.

\section{Evaluation}
To evaluate the effectiveness of each stage of our method, we perform experiments on two commonly used datasets, HumanEva \cite{sigal2010humaneva} and Human3.6M \cite{ionescu2014human3}, and compared with state-of-the-art methods \cite{amin2013multi, belagiannis20143d, elhayek2015efficient, pavlakos2017harvesting, rhodin2016general, sigal2012loose}. Both datasets are collected in a controlled lab environment. HumanEva is composed of 4 different subjects and 6 different motions, while Human3.6M collects sequences from 11 subjects, each performing 15 different motions. To keep compatibility with SMPLify, we also use the first 10 shape parameters in all the experiments, and fine tune all the parameters on the training dataset of HumanEva.

\begin{figure}
\begin{center}
    \begin{subfigure}{0.48\textwidth}
        \scalebox{1.0}[1.0]{\includegraphics[width=0.48\linewidth]{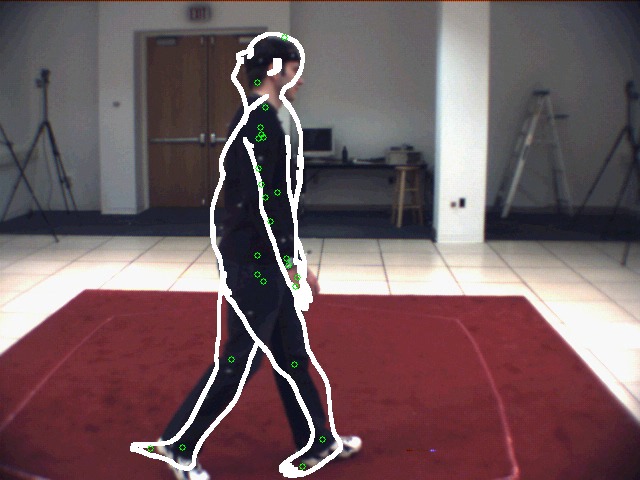}}
        %\caption{}
        \scalebox{1.0}[1.0]{\includegraphics[width=0.48\linewidth]{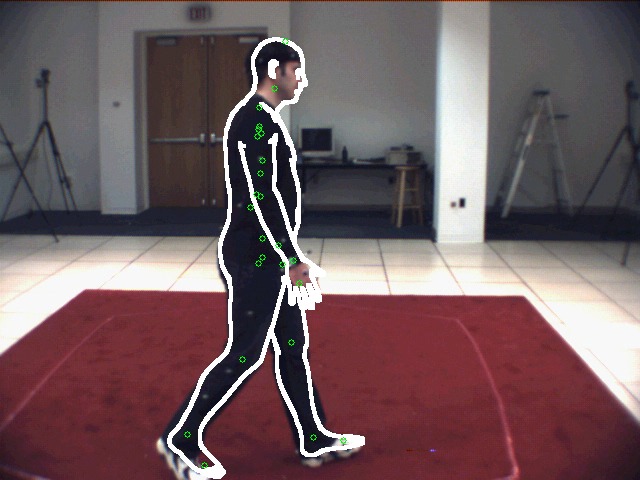}}
        \caption{Correct orientation error}        
    \end{subfigure}   
    \begin{subfigure}{0.48\textwidth}    		
        \scalebox{1.0}[1.0]{\includegraphics[trim=4cm 3cm 7cm 2cm, clip, height=0.50\linewidth]
        {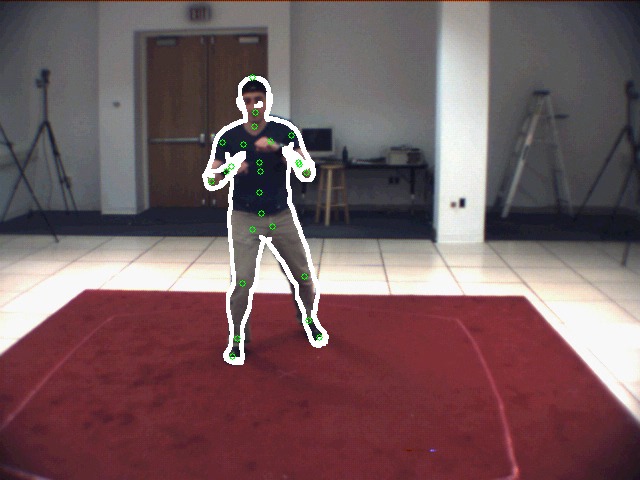}}
        %\caption{}
        \scalebox{1.0}[1.0]{\includegraphics[trim=4cm 3cm 7cm 2cm, clip, height=0.50\linewidth]{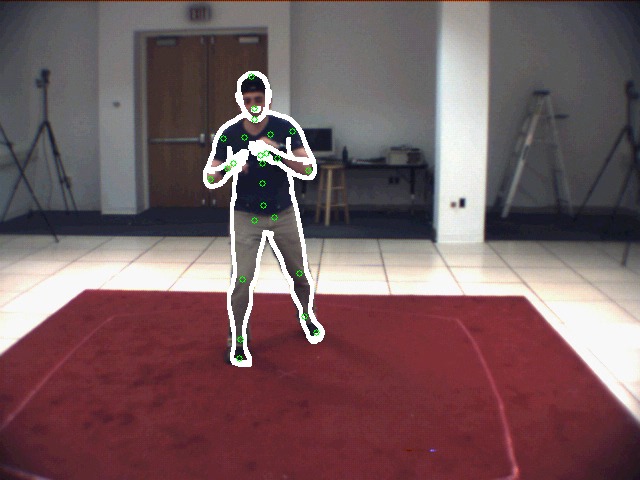}}
        \caption{Better pose}        
    \end{subfigure}          
\caption{MuVS works better than single-view SMPLify. Left column: results of SMPLify, right column: results of MuVS. The white contour represents the projected mesh.}
\label{fig:vssmplify}
\end{center}
\end{figure}

\begin{table}
\begin{center}
%\begin{tabular}{|l c c c c c c c|}
\begin{tabular}{|p{1.7cm} p{0.48cm} p{0.48cm} p{0.48cm} p{0.48cm} p{0.48cm} p{0.48cm} p{0.48cm}|}
\hline
       & \multicolumn{3}{c}{Walking} & \multicolumn{3}{c}{Boxing} & Avg\\
Method & S1 & S2 & S3 & S1 & S2 & S3 & \\
\hline
MuVS\textsuperscript{2} & 18.9 & 19.4 & 20.7 & 19.2 & 13.6 & 20.0 & 18.6 \\
MuVS\textsuperscript{2, S} & 14.6 & 9.6 & 16.6 & 15.1 & 8.5 & 15.8 & 13.4 \\
MuVS\textsuperscript{2, S, T} & 14.1 & 9.3 & 15.9 & 15.0 & 7.4 & 15.1 & 12.8 \\
MuVS\textsuperscript{2, S, T, H} & 12.6 & 5.7 & 6.9 & 12.0 & \textbf{5.4} & \textbf{7.8} & 8.4 \\
MuVS\textsuperscript{3} & 17.6 & 18.6 & 20.6 & 17.2 & 12.3 & 19.4 & 17.6 \\
MuVS\textsuperscript{3, S} & 13.5 & 9.1 & 16.0 & 14.3 & 7.9 & 15.8 & 12.8 \\
MuVS\textsuperscript{3, S, T} & 13.1 & 8.6 & 15.3 & 14.1 & 5.9 & 14.9 & 12.0 \\
MuVS\textsuperscript{3, S, T, H} & \textbf{12.0} & \textbf{5.5} & \textbf{6.4} & \textbf{11.3} & 5.8 & \textbf{7.8} & \textbf{8.1} \\
\hline
\end{tabular}
\end{center}
\caption{Shape estimation error on HumanEva. Error in {\it mm}.}
\label{tab:shape}
\end{table}

\subsection{Ablation study}
To analyze the effect of different parts of our algorithm, firstly we performed various ablation experiments on HumanEva. The estimated 3D locations of joints are compared with that of the ground-truth; unlike other work, no similarity transformation is used unless stated. 
Error is reported in {\it mm} and the results are shown in Table \ref{tab:ablation}. Note that adding the silhouette term in the second stage yields better 3D pose estimation by a large margin, but at the cost of consuming much more running time. % (more than one hour for one unit). 
To make our algorithm comparable with other methods in running-time, we do not use the silhouette term in the temporal fitting stage.  

{\bf Effect of multi-view.} Clearly multiple views can provide more information about the underlying human body. To verify this, we run our algorithm on HumanEva using different numbers of views. As indicated by the result, adding more views consistently improves 3D pose estimation. As shown in Figure \ref{fig:vssmplify} using multiple views helps eliminate incorrectly estimated orientation and improves pose estimation accuracy.

{\bf Effect of silhouette fitting.} Then we conducted experiments to validate the effectiveness of the silhouette term in our method. 
Adding silhouettes consistently improves both 3D pose and shape estimation accuracy. 

{\bf Effect of DCT based temporal prior.} Adding the DCT temporal smoothness term also boosts overall performance. 
As expected its effect diminishes when more views are added, since in this case quite good results can be obtained in the first stage.

{\bf Effect on shape estimation.} To verify the effect of the aforementioned factors on body shape estimation, we run our method on the validation motion sequences of HumanEva, and compare the estimated meshes with those obtained by MoSh \cite{loper2014mosh}. Prior work by Loper et al. \cite{loper2014mosh} shows the generated reference meshes are quite accurate. As evidenced in Table \ref{tab:shape}, adding silhouette information and the DCT temporal prior consistently improves body shape estimation. With 3 views, the average vertext-to-vertex distance is as low as 12 {\it mm} without the silhouette term and around 8 {\it mm} with it.

\subsection{Quantitative comparison}
\textit{HumanEva}: We follow the standard practice of evaluating on the ``Walking" and ``Boxing" sequences of subjects 1, 2 and 3. As in SMPLify \cite{bogo2016keep}, the gender of the subject is assumed known and a gender-specific shape model is used for each motion sequence. The result is shown in Table \ref{tab:com_heva}. Here \textit{General} means the method is trained on the training dataset of HumanEva, instead of separately training the model for each specific subject, which is referred to \textit{Specific}. For the \textit{General} case, we use the joint regressor distributed with SMPL to obtain 3D joints, and directly compare these with the ground truth joint locations. 
For the \textit{Specific} case, we use the joint regressor trained on HumanEva with MoSh, which is provided in SMPLify \cite{bogo2016keep}. 
Then as in \cite{rhodin2016general}, we compute the displacement between the estimated joint location and ground-truth in the first frame, then compensate for this difference in the remaining frames. 

In the \textit{General} case, with only 2 views, our method is more accurate than all the other methods using all 3 views. 
With 3 views we obtain a significant improvement relative to the second best method (55.52 vs 63.25). 
Our method also achieves the lowest error in the \textit{Specific} case. 
Another advantage of our method over the state-of-the-art is that we return a highly realistic body mesh together with skeleton joints. 
Though the method proposed by Rhodin et al. \cite{rhodin2016general} also yields a blob-based 3D mesh, we argue that the underlying SMPL model we use is more realistic. %Furthermore, the mesh variation with pose is inherently considered in SMPL model, which is impossible for \cite{rhodin2016general}. 
A qualitative comparison between our results and those of \cite{rhodin2016general} are shown in Figure \ref{fig:abstract}. 
For more results please refer to our supplementary materials.

\begin{table*}
\begin{center}
\begin{tabular}{|l c c c c c c c c c|}
\hline
      &  & \multicolumn{3}{c}{Walking} & \multicolumn{3}{c}{Boxing} & Mean  & Mean (all)\\
Method & Trained on & S1 & S2 & S3 & S1 & S2 & S3 & &\\
\hline      
Rhodin et al. \cite{rhodin2016general} & \multirow{6}{*}{General} & 74.9 &  &  &  & \textbf{59.7} & & 67.3 & \\
Sigal et al. \cite{sigal2012loose} &  & 66.0 &  &  &  &  &  & 66.0 & \\
Belagiannis et al. \cite{belagiannis20143d} &  & 68.3 &  &  &  & 62.7 &  & 65.5 & \\
Elhayek et al. \cite{elhayek2015efficient} &  & 66.5 &  &  &  & 60.0 &  & 63.25 & \\
MuVS\textsuperscript{2, S, T} & & \textbf{50.14} & 56.11 & 79.55 & 68.96 & 71.73 & 78.45 & \textbf{60.94} & 67.49\\
MuVS\textsuperscript{3, S, T} & & \textbf{43.11} & 53.37 & 73.56 & 64.00 & 67.94 & 71.44 & \textbf{55.52} & 62.23\\
\hline
Amin et al. \cite{amin2013multi} & \multirow{3}{*}{Specific}   & 54.5 &  &  &  & 47.7 &  & 51.10 & \\
Rhodin et al. \cite{rhodin2016general} & & 54.6 &  &  &  & \textbf{35.1} &  & 44.85 & \\
MuVS\textsuperscript{3, S, T} & & \textbf{33.72} & 36.78 & 60.11 & 46.85 & 49.92 & 46.99 & \textbf{41.82} & 45.73 \\
\hline
\end{tabular}
\end{center}
\caption{Quantitative comparison on HumanEva. 3D joint errors in {\it mm}.}
\label{tab:com_heva}
\end{table*}

\textit{Human3.6M}: To further validate the generality and usefulness of MuVS, we also evaluate it on Human3.6M \cite{ionescu2014human3}. Human3.6M is the largest public dataset for pose estimation, composed of a wide range of motion types, some of them being very challenging. We use the same parameters trained on HumanEva, then apply MuVS on all the 4 views of subjects S9 and S11. We compare it with SMPLify \cite{bogo2016keep} and other state-of-the-art multi-view pose estimation methods \cite{pavlakos2017harvesting}. The result is shown in Table \ref{tab:com_h36_1}. 
The multi-view version is significantly more accurate than SMPLIfy and our 3D joint estimation accuracy is quite close to that of \cite{pavlakos2017harvesting}, which is concurrent with our work. 
While they only focus on 3D joint estimation,  we  address 3D pose and shape estimation at the same time. Our method not only returns 3D joint estimates, but also a realistic body shape model that  is faithful to the subjects and
which is ready for later modification and animation. 
%This is not easily achievable for the method proposed in \cite{pavlakos2017harvesting}.

\begin{table*}
\begin{center}
{\small
\begin{tabular}{|l c c c c c c c c c|}
\hline
  & Directions & Discussion & Eating & Greeting & Phoning & Photo & Posing & Purchases & Sit \\
\hline
SMPLify \cite{bogo2016keep} & 62.0 & 60.2 & 67.8 & 76.5 & 92.1 & 77.0 & 73.0 & 75.3 & 100.3 \\
MuVS\textsuperscript{4, S, T, Sim} & \textbf{35.05} & \textbf{39.22} & \textbf{38.59} & \textbf{37.35} & \textbf{59.16} & \textbf{46.07} & \textbf{40.52} & \textbf{38.47} & \textbf{60.07} \\
\hline
Tekin et al. \cite{tekin2015direct} & 102.41 & 147.72 & 88.83 & 125.28 & 118.02 & 182.73 & 112.38 & 129.17 & 138.89 \\
MuVS\textsuperscript{4, S, T} & 44.32 & \textbf{46.99} & 51.75 & 44.99 & 67.68 & 54.56 & 49.25 & \textbf{48.90} & \textbf{72.82} \\
Pavlakos et al.\cite{pavlakos2017harvesting} & \textbf{41.18} & 49.19 & \textbf{42.79} & \textbf{43.44} & \textbf{55.62} & \textbf{46.91} & \textbf{40.33} & 63.68 & 97.56 \\
\hline
  & SitDown & Smoking & Waiting & WalkDog & Walk & WalkTogether & Mean & Median & \\
\hline
SMPLify \cite{bogo2016keep} & 137.3 & 83.4 & 77.3 & 79.7 & 86.8 & 81.7 & 82.3 & 69.3 & \\
MuVS\textsuperscript{4, S, T, Sim} & \textbf{69.70} & \textbf{56.24} & \textbf{67.91} & \textbf{46.91} & \textbf{38.00} & \textbf{33.15} & \textbf{47.09} &  \textbf{40.52} & \\
\hline
Tekin et al. \cite{tekin2015direct} & 224.9 & 118.42 & 138.75 & 126.29 & 55.07 & 65.76 & 124.97 & 125.28 & \\
MuVS\textsuperscript{4, S, T} & \textbf{76.51} & 63.70 & 116.24 & 55.44 & 42.94 & \textbf{37.24} & 58.22 & 51.75 & \\
Pavlakos et al.\cite{pavlakos2017harvesting} & 119.90 & \textbf{52.12} & \textbf{42.68} & \textbf{51.93} & \textbf{41.79} & \textbf{39.37} & \textbf{56.89} & \textbf{46.91} & \\
\hline
\end{tabular}
}
\end{center}
\caption{Quantitative comparison with SMPLify, the methods of Tekin et al. \cite{tekin2015direct} and Pavlakos et al. \cite{pavlakos2017harvesting} on H3.6M dataset in {\it mm}. The accuracy of our method is comparable with that of the recent method proposed in \cite{pavlakos2017harvesting}.}
\label{tab:com_h36_1}
\end{table*}

\begin{table}
\begin{center}
%\begin{tabular}{|l c c c c c c c|}
\begin{tabular}{|p{1.7cm} p{0.48cm} p{0.48cm} p{0.48cm} p{0.48cm} p{0.48cm} p{0.48cm} p{0.48cm}|}
\hline
       & \multicolumn{3}{c}{Walking} & \multicolumn{3}{c}{Boxing} & Avg\\
Method & S1 & S2 & S3 & S1 & S2 & S3 & \\
\hline
SMPLify\cite{bogo2016keep} & 73.3 & 59.0 & 99.4 & 82.1 & 79.2 & 87.2 & 79.9 \\
MuVS\textsuperscript{1, S, Sim} & 51.3 & 48.2 & \textbf{80.9} & 68.4 & 80.7 & 88.7 & 69.7\\
MuVS\textsuperscript{1, S, T, Sim} & \textbf{51.2} & \textbf{48.1} & 81.6 & \textbf{61.5} & \textbf{78.3} & \textbf{82.6} & \textbf{67.2} \\
\hline
\end{tabular}
\end{center}
\caption{Comparison with SMPLify on monocular videos from HumanEva in mm. Here Sim means using Procrustes analysis per frame, as with SMPLify.}
\label{tab:monocular}
\end{table}

\begin{figure*}
\begin{center}
    %\begin{subfigure}{\textwidth}
        \scalebox{1.0}[1.0]{\includegraphics[width=0.19\linewidth]{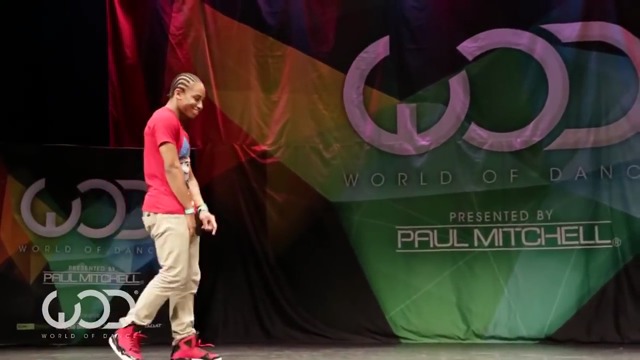}}
        %\caption{}
        \scalebox{1.0}[1.0]{\includegraphics[width=0.19\linewidth]{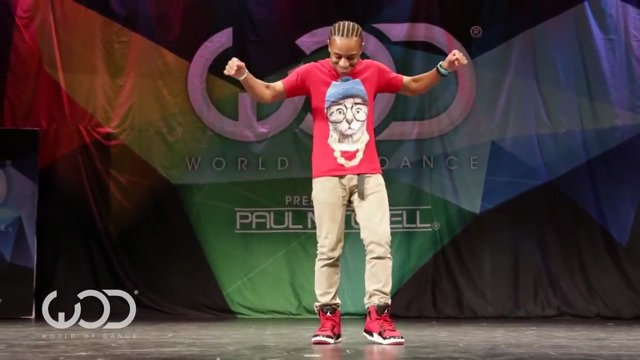}} 
        \scalebox{1.0}[1.0]{\includegraphics[width=0.19\linewidth]{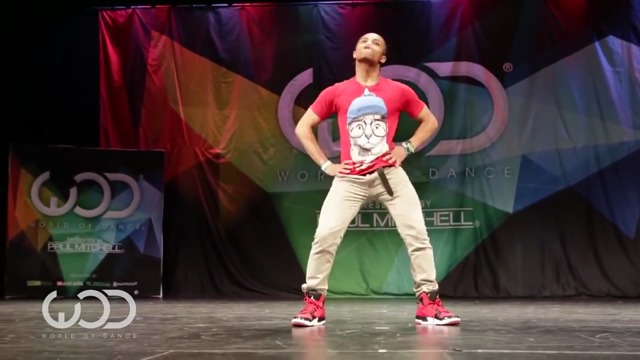}}
        \scalebox{1.0}[1.0]{\includegraphics[width=0.19\linewidth]{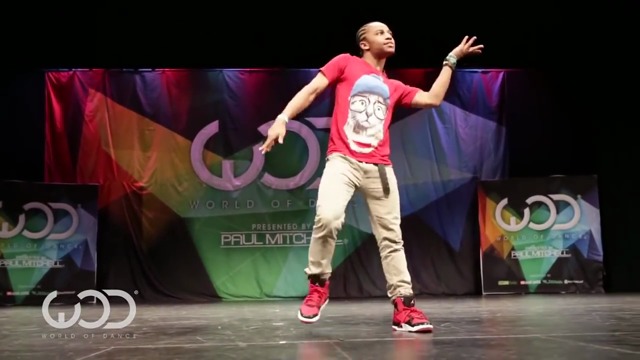}}
        \scalebox{1.0}[1.0]{\includegraphics[width=0.19\linewidth]{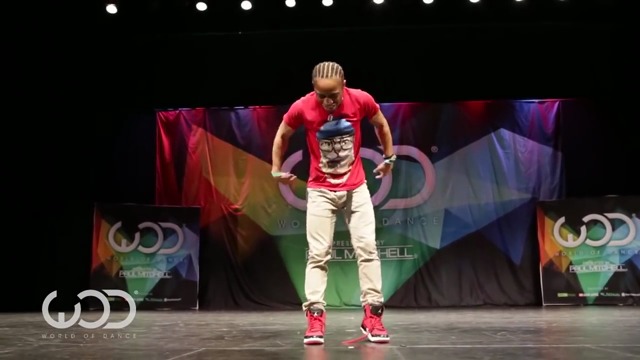}} 
        %\caption{Correct orientation error}        
    %\end{subfigure}    \\
    %\begin{subfigure}{\textwidth}
        \\
        \scalebox{1.0}[1.0]{\includegraphics[width=0.19\linewidth]{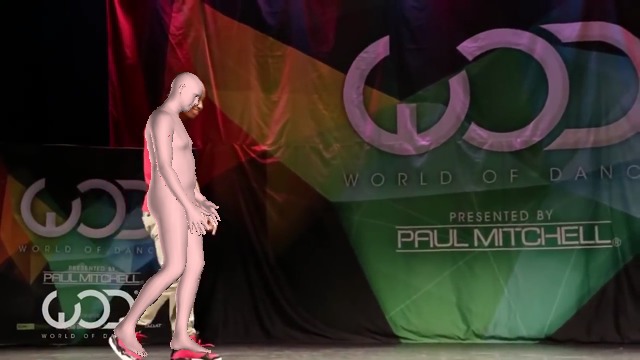}}
        %\caption{}
        \scalebox{1.0}[1.0]{\includegraphics[width=0.19\linewidth]{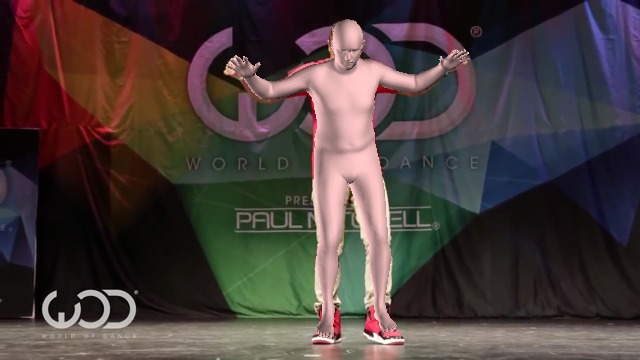}} 
        \scalebox{1.0}[1.0]{\includegraphics[width=0.19\linewidth]{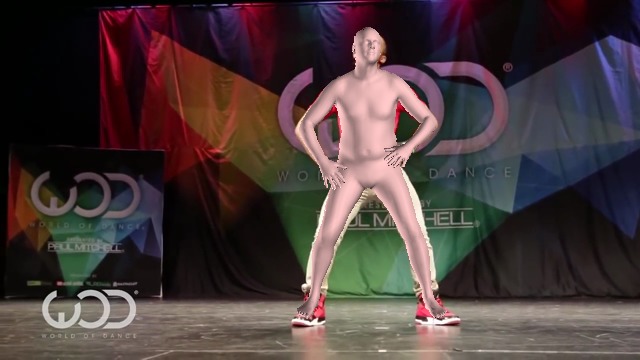}}
        \scalebox{1.0}[1.0]{\includegraphics[width=0.19\linewidth]{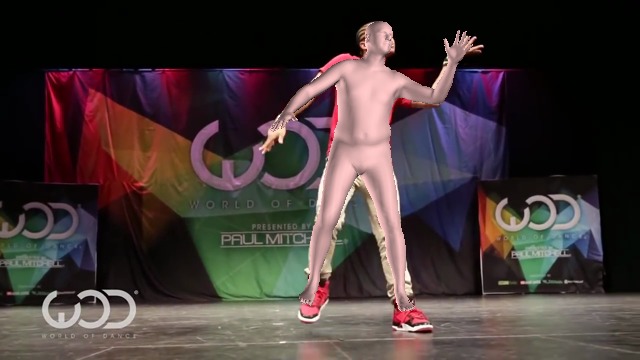}}
        \scalebox{1.0}[1.0]{\includegraphics[width=0.19\linewidth]{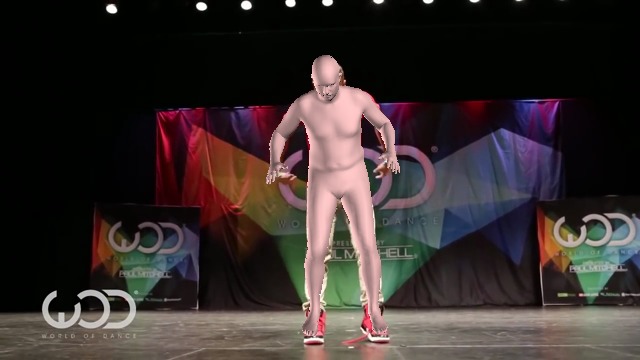}} 
        %\caption{Better pose}        

        \scalebox{1.0}[1.0]{\includegraphics[width=0.19\linewidth]{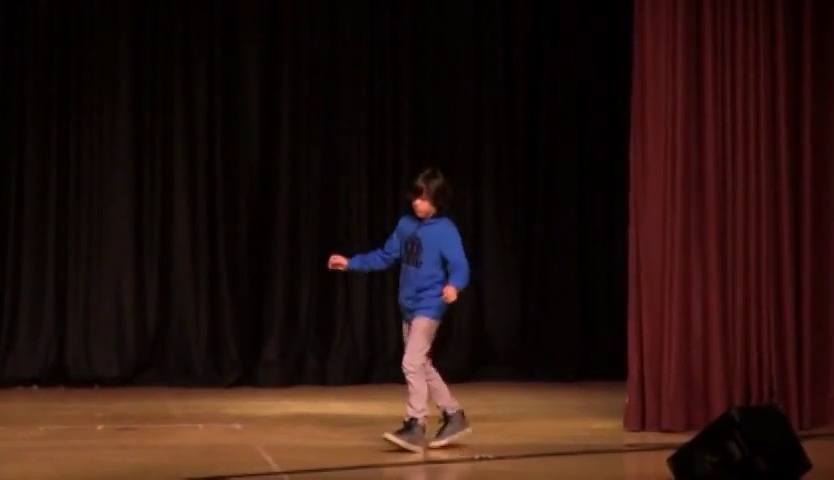}}
        %\caption{}
        \scalebox{1.0}[1.0]{\includegraphics[width=0.19\linewidth]{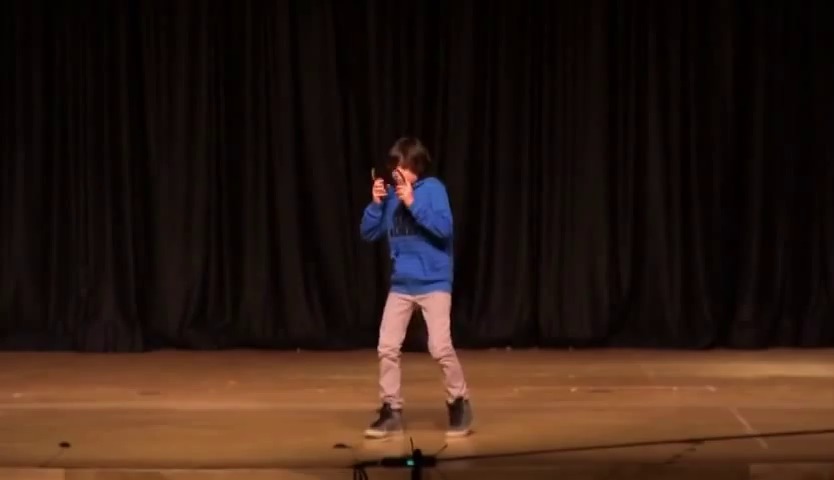}} 
        \scalebox{1.0}[1.0]{\includegraphics[width=0.19\linewidth]{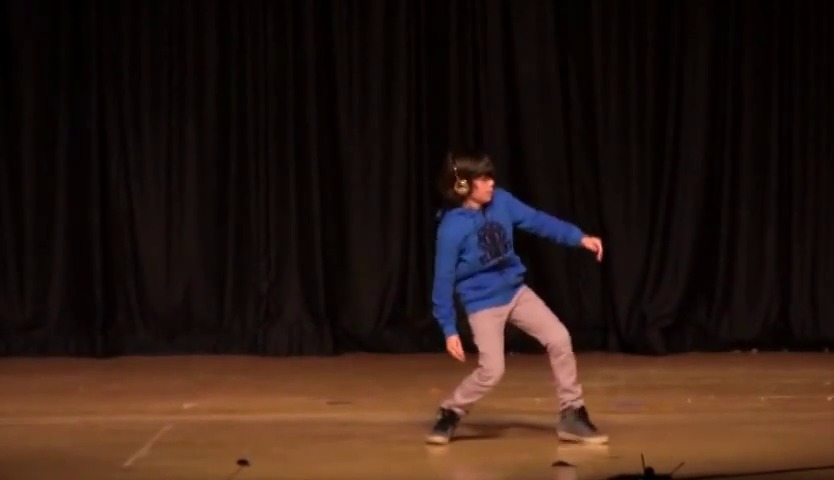}}
        \scalebox{1.0}[1.0]{\includegraphics[width=0.19\linewidth]{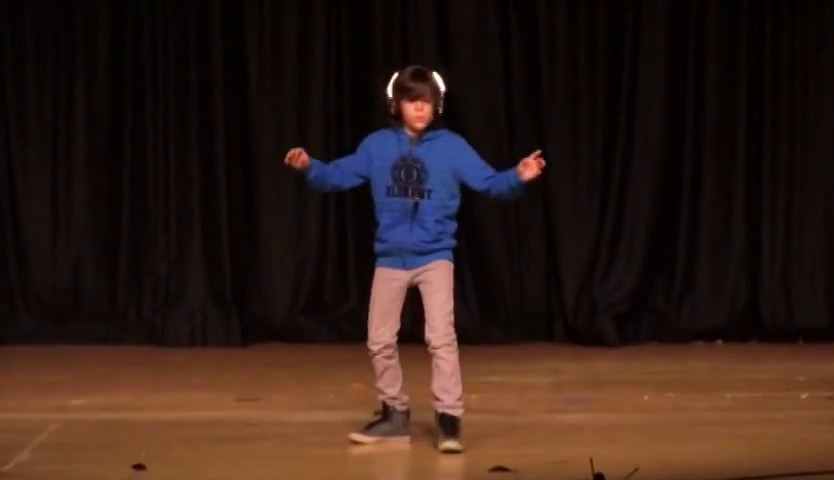}}
        \scalebox{1.0}[1.0]{\includegraphics[width=0.19\linewidth]{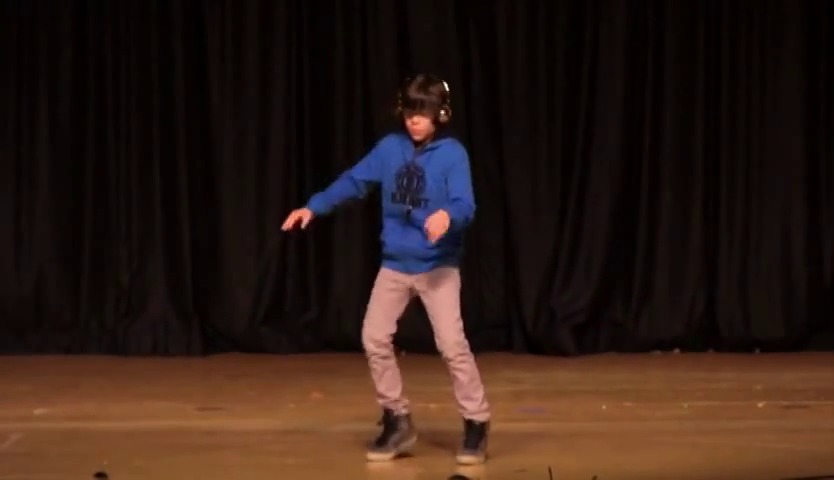}} 
        \scalebox{1.0}[1.0]{\includegraphics[width=0.19\linewidth]{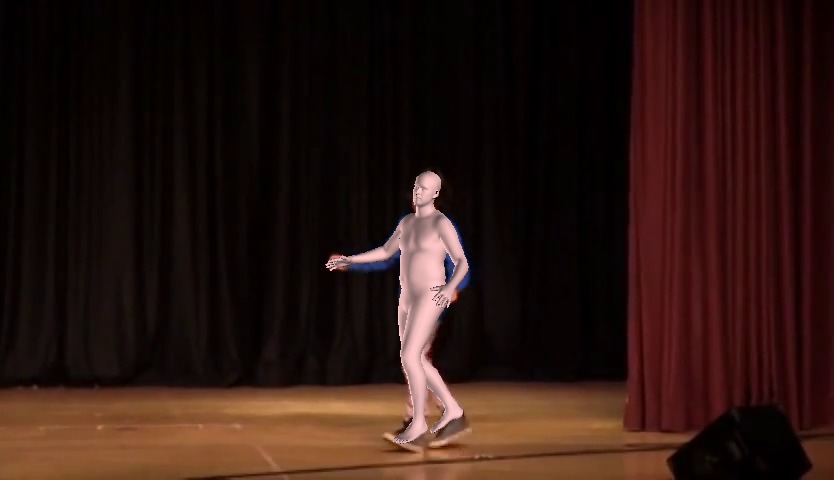}}
        %\caption{}
        \scalebox{1.0}[1.0]{\includegraphics[width=0.19\linewidth]{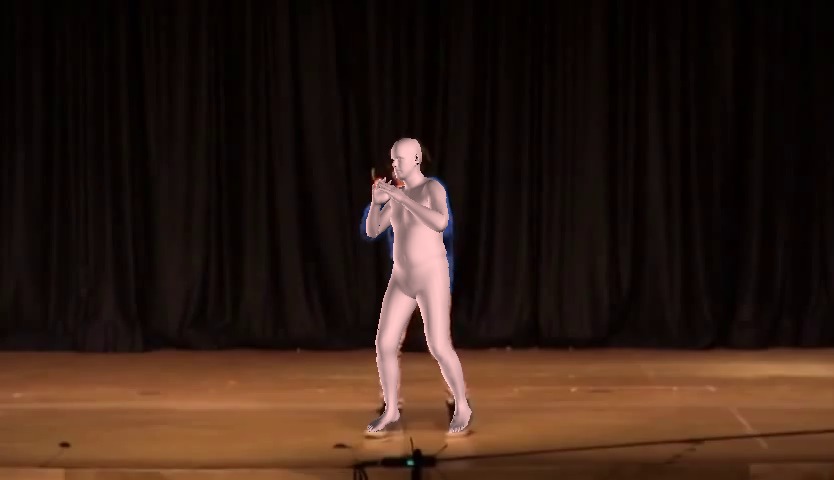}} 
        \scalebox{1.0}[1.0]{\includegraphics[width=0.19\linewidth]{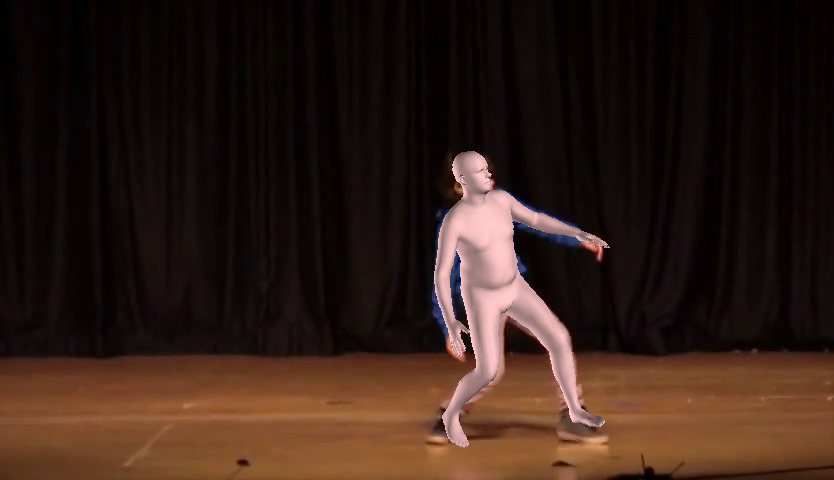}}
        \scalebox{1.0}[1.0]{\includegraphics[width=0.19\linewidth]{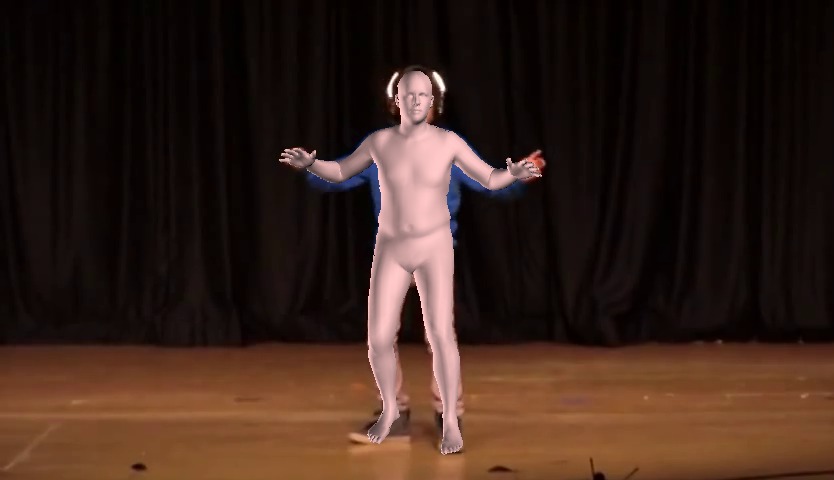}}
        \scalebox{1.0}[1.0]{\includegraphics[width=0.19\linewidth]{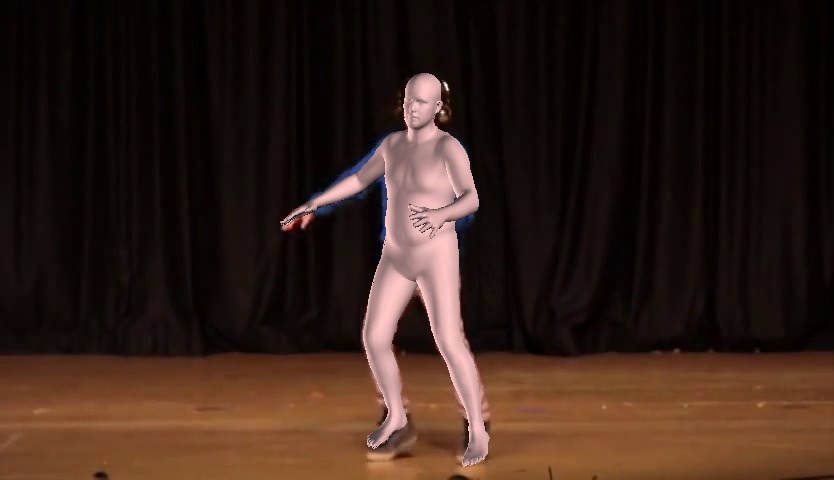}}

    %\end{subfigure}          
\caption{Monocular pose estimation results on videos downloaded from YouTube.}
\label{fig:monocularextra}
\end{center}
\end{figure*}

%\begin{table}
%\begin{center}
%\begin{tabular}{|l c c c|}
%\begin{tabular}{|p{1.7cm} p{0.6cm} p{0.6cm} p{0.6cm}|}
%\hline
%Method  & Precision & Recall & F-Value\\
%\hline
%Helge et al. \cite{rhodin2016general} & & & \\
%MuVS & & & \\
%\hline
%\end{tabular}
%\end{center}
%\caption{Shape comparison on HumanEva.}
%\label{tab:shape}
%\end{table}

\begin{figure*}
\begin{center}
        \includegraphics[width=0.97\linewidth]{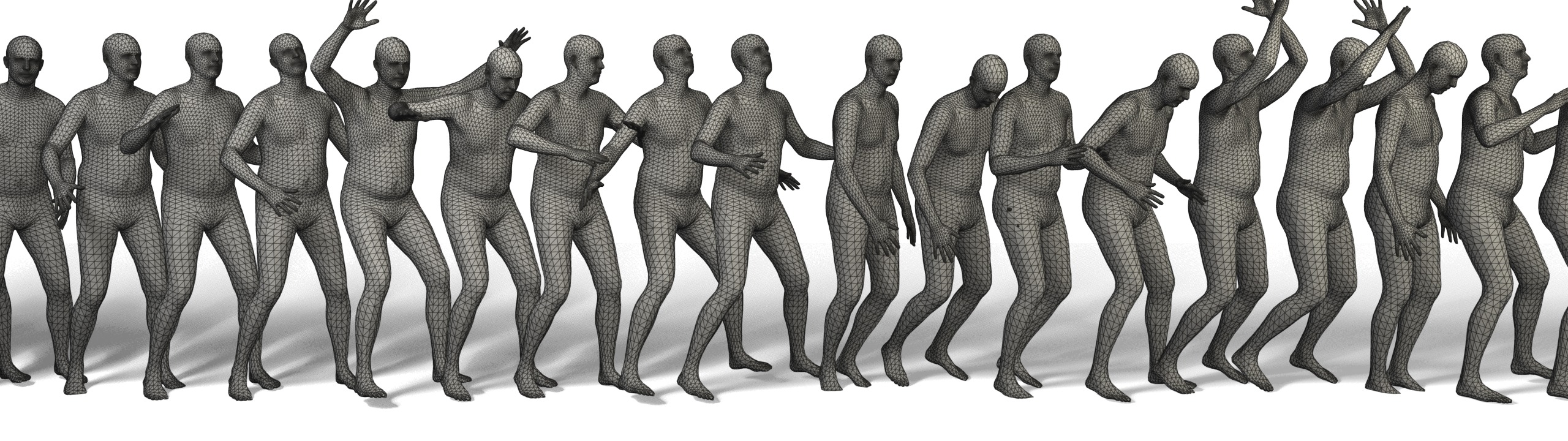}
\caption{Demonstration of generated meshes for a monocular motion sequence from YouTube.}
\label{fig:monocular}
\end{center}
\end{figure*}

\section{Pose and Shape from Monocular Video}
Though we focus on multi-view pose and shape estimation, our method can be applied to monocular video sequences without large modifications, while still being fully automatic. Note manually initialized pose is required for the method in \cite{rhodin2016general} to work on monocular data. 

We compare our method with SMPLify on the first camera view of HumanEva, and the result is shown in Table \ref{tab:monocular}. 
Of course given only a single video, it is hard to apply the DCT constraint in depth, since we do not have any trustable evidence in that dimension. 
Empirically we find our method can still return quite promising results when the performer does not move much in depth. 
For the videos where no camera information is provided, we manually set the focal length and other imaging parameters to some common value as done in \cite{bogo2016keep}. We qualitatively evaluate our method on some videos downloaded from YouTube, and show the results for specific frames in Figure \ref{fig:monocularextra}. Figure \ref{fig:monocular} shows the reconstructed mesh sequence of one of the videos. For the full video, please refer to our supplementary materials.

\section{Conclusion and Future Work}
In this paper we present a new marker-less motion capture system, MuVS, that extends SMPLify in a principled and straightforward way. Our method computes relatively accurate 3D pose and also returns a realistic and faithful human body mesh.
Unlike previous work that assumes known silhouettes, needs user intervention, or limits the user motion, our algorithm works for general activities seen in daily life. 
Evaluation on public benchmarks validates the effectiveness and generality of our method. 
Additionally we apply the approach to monocular video sequences, and achieve promising results. 

Future work will address more complex scenarios, like cluttered backgrounds, multiple people, and extreme poses. 
A key direction to make the method practical is to reduce the computational costs.
%We will also try to speed up our algorithm to make it more applicable for real-world applications. 
Finally, other body parts, like faces, hands and feet could be easily combined into our model. 

\textbf{Acknowledgements}. We thank the anonymous reviewers for their insightful comments, Naureen Mahmood for help with the figures, Ahmed Osman for voice recording, and Tianye Li and Timo Bolkart for proofreading.

\end{document}